\documentclass{article}

\usepackage{arxiv}

\usepackage[utf8]{inputenc} 
\usepackage[T1]{fontenc}    
\usepackage{hyperref}       
\usepackage{amsfonts}       
\usepackage{nicefrac}       
\usepackage{microtype}      
\usepackage{lipsum}
\usepackage{pdfpages}

\usepackage{subcaption}
\usepackage{booktabs}
\usepackage{verbatim}
\usepackage{amsmath}
\usepackage[round]{natbib}
\def\UrlFont{\rm}  
\usepackage{graphicx}  
\title{Distilling BERT into Simple Neural Networks with Unlabeled Transfer Data\thanks{Multilingual version of this work appears at ACL~\citep{mukherjee-hassan-awadallah-2020-xtremedistil}.}}
\author{
  Subhabrata Mukherjee \\
  Microsoft Research\\
  Redmond, WA \\
  \texttt{subhabrata.mukherjee@microsoft.com} \\
   \And
 Ahmed Hassan Awadallah \\
  Microsoft Research\\
  Redmond, WA\\
  \texttt{hassanam@microsoft.com} \\
}

\begin{document}

\maketitle
   
\begin{abstract}

Recent advances in pre-training huge models on large amounts of text through self supervision have obtained state-of-the-art results in various natural language processing tasks. However, these huge and expensive models are difficult to use in practise for downstream tasks. Some recent efforts~\citep{DBLP:journals/corr/abs-1903-12136,sun2019patient,sanh2019,turc2019wellread} use knowledge distillation to compress these models. However, we see a gap between the performance of the smaller student models as compared to that of the large teacher. In this work, we leverage large amounts of in-domain unlabeled transfer data in addition to a limited amount of labeled training instances to bridge this gap. We show that simple RNN based student models even with hard distillation can perform at par with the huge teachers given the transfer set. The student performance can be further improved with soft distillation and leveraging teacher intermediate representations. We show that our student models can compress the huge teacher by up to $26x$ while still matching or even marginally exceeding the teacher performance in low-resource settings with small amount of labeled data. Additionally, for the multilingual extension of this work with XtremeDistil~\citep{mukherjee-hassan-awadallah-2020-xtremedistil}\footnote{Code available at: \url{https://aka.ms/XtremeDistil}}, we demonstrate massive distillation of multilingual BERT-like teacher models by upto 35x in terms of parameter compression and 51x in terms of latency speedup for batch inference while retaining 95\% of its F1-score for NER over 41 languages.

\end{abstract}

\section{Introduction}

\noindent{\bf Motivation:} Deep neural networks are the state-of-the-art for various natural language processing applications like text classification, named entity recognition, question-answering, etc. However, one of the biggest challenges facing them is the lack of labeled data to train these complex networks. Recent advances in pre-training help close this gap. In this, deep and large neural networks are trained on millions to billions of documents in a self-supervised fashion to obtain general purpose language representations. Table~\ref{tab:size} shows a comparison of the model size of the state-of-the-art pre-trained language models.

\begin{table}[h]
    \centering
    \begin{tabular}{lrr}
    \toprule
Pre-trained Language Models	& \#Training & \#Model\\
	& Words	& Parameters\\
	\midrule
CoVe~\citep{DBLP:conf/nips/McCannBXS17}	& 20M &	30M\\
ELMO~\citep{DBLP:conf/naacl/PetersNIGCLZ18}	& 800M &	90M\\
GPT~\citep{radford2018improving}	& 800M &	100M\\
BERT~\citep{DBLP:conf/naacl/DevlinCLT19} &	3.3B &	340M\\
XLNet~\citep{DBLP:journals/corr/abs-1906-08237} &	32.89B &	340M\\
GPT 2~\citep{radford2019} &	8B &	1.5B\\
\bottomrule
    \end{tabular}
    \caption{Pre-trained language model complexity.}
    \label{tab:size}
\end{table}

A significant challenge facing many practitioners is how to deploy these huge models in practice. Although these models are trained offline, during prediction we still need to traverse the deep neural network architecture stack involving a large number of parameters that significantly increases latency and memory requirements.

\noindent {\bf State-of-the-art:} Prior works, especially in the computer vision area, addressed this issue using a framework called student teacher distillation. In this, first, a large, cumbersome and accurate neural network (called the {\em teacher}) is trained to extract structure from the data. 
The second phase is called {\em distillation} where the knowledge from the teacher is transferred to a smaller model (called the {\em student}) that is easier to deploy. Prior works have explored techniques like training the student on soft targets (called the {\em logits}) from the teacher~\citep{DBLP:conf/nips/BaC14}, adjusting the temperature of the softmax~\citep{DBLP:journals/corr/HintonVD15} or training thin and deep networks~\citep{DBLP:journals/corr/RomeroBKCGB14}. In contrast to the vision community, there has been limited exploration of such techniques in the NLP community.

\noindent{\bf Concurrent Works:} \citep{DBLP:journals/corr/abs-1904-09482} distil knowledge from several huge multi-task learners into a single huge learner. In this work, we focus on distilling knowledge from a huge learner into a simple learner. Similar approach has also been studied in~\citep{DBLP:journals/corr/abs-1903-12136} with rule based data augmentation using various heuristics when no unlabeled data is available. \citep{sun2019patient} propose a task-specific knowledge distillation approach in which they use a subset of the layers of the huge pre-trained language model to construct a shallower student model. A similar approach is also used in \citep{sanh2019}. Since these works initialize their student models with a subset of the learned representations of the transformer, the distilled student is constrained by the architecture of the teacher. \citep{turc2019wellread} address this shortcoming by pre-training a shallow transformer with a different architecture than the teacher on unlabeled data and then performing the distillation. However, pre-training is time consuming and resource extensive. Table~\ref{tab:distil-comparison} shows an overview of the different distillation methods. We compare them on the following aspects:(i) the architecture of the student model, (ii) whether the distillation process employs pre-training, (iii) whether the student model is constrained by the architecture of the teacher, (iv) different teacher model sizes, (v) compression given as the ratio of the number of parameters in the teacher model to that in the student model and (vi) the performance gap between the teacher performance and the student performance aggregated over all the datasets used in the respective experiments. However, please note that the different works use different subsets of the data, and, therefore not readily comparable on the performance gap for different settings.

\begin{table}[h]
\small
    \centering
    \begin{tabular}{lccccrr}
    \toprule
Work & Architecture & Pre-training & Teacher-constrained & Teacher & Compression & Performance Gap\\\midrule
\citep{DBLP:journals/corr/abs-1903-12136} & BiLSTM & No & No & BERT Large & $349x$ & $9\%$\\
 &  &  &  & BERT Base & $114x$ & $7\%$\\\midrule
\citep{sanh2019} & Transformer & Yes & Yes & BERT Base & $1.7x$ & $5\%$\\\midrule
\citep{sun2019patient} & Transformer & No & Yes & BERT Base & $1.7x$ &  $2.3\%$\\
                                    & & &  &  BERT Base & $2.4x$ & $7\%$\\
                                 &   &  &  & BERT Large & $5x$ & $9\%$\\\midrule
\citep{turc2019wellread} & Transformer & Yes & No & BERT Base & $1.7x$ & < $1\%$\\
 &  &  &  & BERT Large & $8x$ &  $2.5\%$\\
 &  &  &  & BERT Large & $12x$ &  $3.2\%$\\
 &  &  &  & BERT Large & $30x$ &  $4.2\%$\\
  &  &  &  & BERT Large & $77x$ & $8\%$\\
\midrule
Ours (HR) & BiLSTM & No & No & BERT Base & $9x$ & {\em M} \\
Ours (HR) &  &  &  & BERT Large & $26x$ & < $1\%$ \\
Ours (LR) &  &  &  & BERT Large & $26x$ & {\em M} \\

\bottomrule 
    \end{tabular}
    \caption{Comparison of pre-trained language model distillation methods. A higher compression factor is better, whereas lower performance gap is better. {\em M} indicates a matching performance with the student marginally better than the teacher. HR indicates high resource setting with all training labels; LR indicates low-resource with $100$ training labels per class.}
    \label{tab:distil-comparison}
\end{table}

\begin{figure}
\centering
\begin{subfigure}{.5\textwidth}
  \centering
  \includegraphics[width=0.97\linewidth]{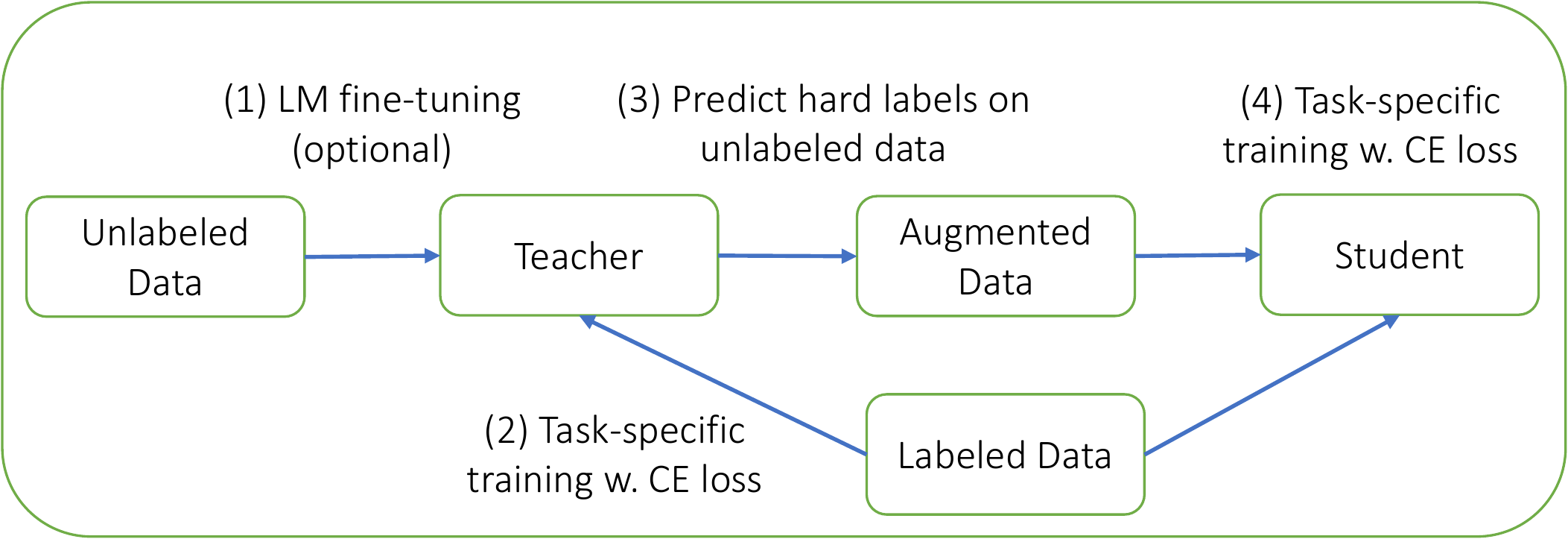}
  \caption{Hard distillation.}
  \label{fig:sub1}
\end{subfigure}%
\begin{subfigure}{.5\textwidth}
  \centering
  \includegraphics[width=0.97\linewidth]{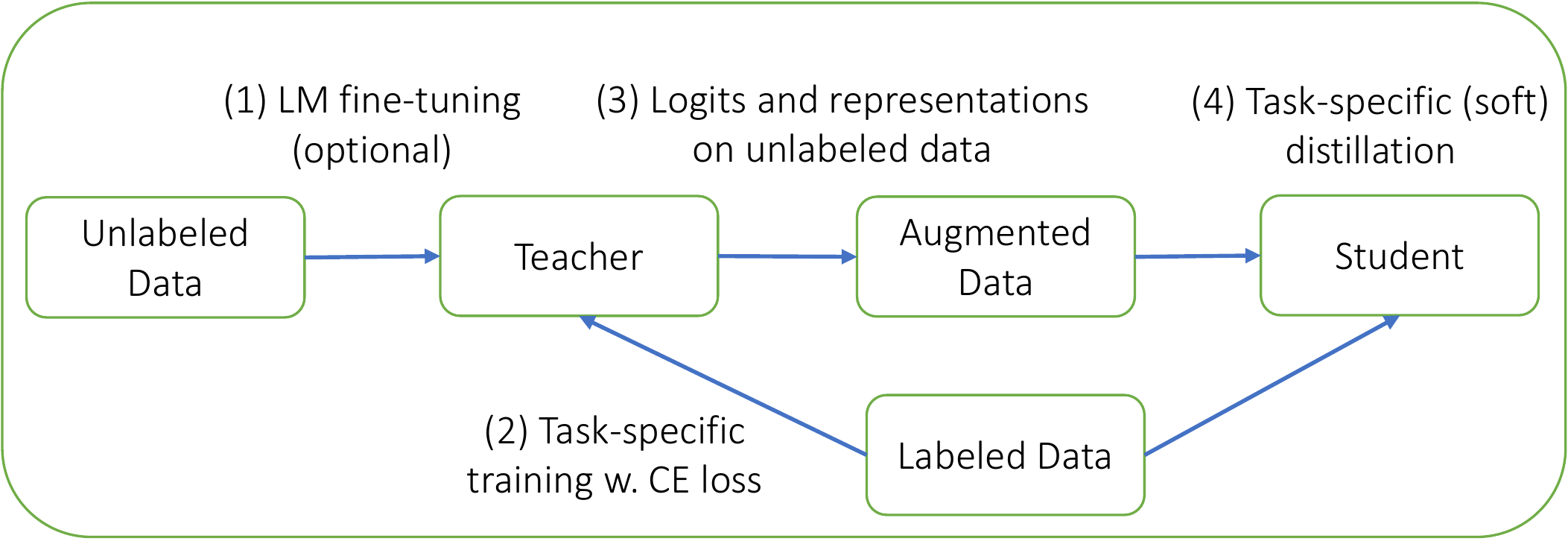}
  \caption{Soft distillation.}
  \label{fig:sub1}
\end{subfigure}%
  \caption{Overview of different distillation methods.}
  \label{fig:overview}
\end{figure}

{\noindent \bf Overview of our work:} In this work, we study knowledge distillation in the presence of a limited amount of labeled training data and a large amount of unlabeled transfer data. We study several aspects of the distillation process like the distillation objective, how to harness the teacher representations, the training schedules, the impact of the amount of labeled training data, the size of the teacher, to name a few. In contrast to prior works, we show that RNN encoders like BiLSTMs as students can perform at par with the large pre-trained language models with greatly reduced parameter space ($9x$-$26x$) given a sufficiently accurate teacher and a large amount of unlabeled data with distillation. We explore two distillation techniques. The first one is {\em hard distillation}, where we use the fine-tuned teacher to auto-annotate a large amount of unlabeled data with hard labels and then use the augmented data to train the student with supervised cross-entropy loss. We show that this simple method works surprisingly well given very confident predictions from the transformers. In the second {\em soft distillation} technique, we use {\em logits} and {\em internal representations} from the transformer generated on the unlabeled data to train the student model with different training schedules to optimize for different loss functions. We show that soft distillation can further regularize these models and improve performance. Also, note that our approach is not constrained by the size or embedding dimension of the teacher although we leverage some of its internal representations. Figure~\ref{fig:overview} shows an overview of the hard and soft distillation methods. 

\noindent{\bf Our contributions:} We perform an extensive study of knowledge distillation from large pre-trained language models. 
Our study focuses on the following dimensions:
\begin{itemize}
    \item {\bf Models and Features:} We explore different representations and features to distil knowledge from the teacher as well as different distillation objectives (soft logits vs. hard targets). We leverage internal representations from the teacher for distillation such that our approach is not constrained by the teacher embedding dimension.
    \item {\bf Training:} We explore several optimization strategies for training and show that the training schedule has an impact on the distillation performance and show one of them to consistently perform well. 
    \item {\bf Experiments:} We perform extensive experiments on four large-scale datasets with varying number of classes and number of labeled training instances per class. We analyze several scenarios where distillation can provide added value not only in terms of compression but also on improving the performance. We empirically show why soft distillation is difficult for deep pre-trained language models that produce very confident predictions and offer limited uncertainties to explore. On the same note, we show that simple hard distillation techniques perform extremely well given a large unlabeled transfer set.
\end{itemize}

\section{Problem Description}

Consider $D_l = \{x_l, y_l\}$ to be a set of $n$ labeled instances with $X=\{x_l\}$ denoting the instances and $Y=\{y_l\}$ the corresponding labels. Consider $D_u = \{x_u\}$ to be a transfer set of $N$ unlabeled instances from the same domain. We assume a realistic setting where there are few labeled instances and a large number of unlabeled instances i.e. $n << N$.

Now, given a teacher model $\mathcal{T}(\theta^t)$ with $\theta^t$ being the set of trainable parameters, we want to train a student model $\mathcal{S}(\theta^s)$ with parameters $\theta^s$ such that $|\theta^s| << |\theta^t|$ and the student model is comparable in performance to that of the teacher based on some evaluation metric.

In the following section, the superscript `t' always represents the teacher and `s' denotes the student.

\noindent{\bf Input Representation} We tokenize all the text sequences using Wordpiece tokenization~\citep{DBLP:journals/corr/WuSCLNMKCGMKSJL16}. This considers a fixed vocabulary size (e.g., $30k$ tokens) and segments words into wordpieces such that every piece is present in the vocabulary. A special symbol `\#\#' is used to mark the split of a word into its pieces. Following prior works~\citep{DBLP:conf/naacl/DevlinCLT19}, we add the special symbols ``[CLS]" and ``[SEP]" to mark the beginning and end of a text sequence respectively. For example, a given sequence ``mobilenote to ms. jacobson and ms. ferrer" is now tokenized as ``[CLS] mobile \#\#note to ms . jacobs \#\#on and ms . ferrer [SEP]".

\section{Models}

\subsection{The Student}
\label{subsec:student}
Our student neural network architecture comprises of three layers stacked on top of each other. The first one is the word embedding layer that generates a $K$ dimensional representation for each token. 

In order to capture the sequential information in the tokens, we use Bidirectional Long Short Term Memory Networks (BiLSTM). Given a sequence of $t$ tokens, a BiLSTM computes a set of $T$ vectors $h_{t}$ as the concatenation of the states generated by a forward and backward LSTM. Assuming the number of hidden units in the LSTM to be $L$, each hidden state $h_t$ is of dimension $2L$.

The last hidden state $h_t$ of the input sequence can be used as an aggregate representation for the entire sequence. However, some prior works~\citep{DBLP:conf/emnlp/ConneauKSBB17} have shown that for longer sequences, the last state may not be sufficient to encapsulate all the information. Therefore, we adopt a {\em max pooling} mechanism~\citep{DBLP:conf/icml/CollobertW08}. This selects the maximum feature value $z^{(s)}_{l}$ over each temporal dimension of the input sequence.

\begin{align*}
\overrightarrow{h_t} &= \overrightarrow{LSTM}(w_1, w_2, \cdots w_t)\\
\overleftarrow{h_t} &= \overleftarrow{LSTM}(w_1, w_2, \cdots w_t)\\
h_t &= [\overrightarrow{h_t}; \overleftarrow{h_t}]\\
z^{(s)} &= \{max_t\ h_{t,l}\}
\end{align*}

Now, given the student representation $z^{(s)}(x; \theta_s)=\{z^{(s)}_l(x; \theta_s)\}$ for an instance $x$, the classification probabilities can be computed as:

\begin{equation}
    \label{eq:0}
p^{(s)}(x) = softmax(z^{(s)}(x; \theta_s) \cdot W^{(s)})
\end{equation}
\noindent where $W^{(s)} \in R^{H.C}$ and $C$ is the number of classes. 

We want to train the student neural network end-to-end by solving the following optimization equation:

\begin{equation}
\label{eq:1}
    \min_{\theta^s, W^s} \mathbb{E}_{(x_l,y_l) \in \mathcal{D}_l} \mathcal{L}_{CE}(y_l, p^s(x_l;\theta_s,W^s))
\end{equation}

\noindent where $\mathcal{L_{CE}}$ denotes the cross entropy loss given by:

\begin{equation}
\label{eq:2}
    \mathcal{L_{CE}} = \sum_{x_l, y_l \in D_l} \sum_c y_{l,c}\  log\ p_c^{(s)}(x_l)
\end{equation}

In absence of any distillation, this is the standard objective for classification and constitutes one of our baselines.

\subsection{The Teacher}

    Recently, there has been a paradigm shift in the field of Natural Language Processing with the emergence of deep pre-trained language models like ELMO~\citep{DBLP:conf/naacl/PetersNIGCLZ18}, BERT~\citep{DBLP:conf/naacl/DevlinCLT19} and GPT~\citep{radford2018improving,radford2019} that have shown state-of-the-art performance for several tasks. These models are trained by self supervision using objectives like next word prediction and masked language model objectives on massive corpora. We adopt one of these deep pre-trained language models as our teacher.

\subsubsection{Fine-tuning the Teacher}
The pre-trained language models are trained for general language model objectives. In order to adapt these models for the given task, we fine-tune them on the labeled data $D_l=\{x_l. y_l\}$. 

Given the wordpiece tokenized input sequence $x$, we use the last hidden state representation $z^{(t)}(x; \theta_t)$ from the pre-trained language model corresponding to the first input ``[CLS]" token as an aggregate representation. Similar to the student model, we can add a softmax classification layer with parameters $W^{(t)} \in R^{H \cdot C}$ where $C$ is the number of classes. The learning objective is to minimize the classification loss on $p^{(t)}(x) = softmax(z^{(t)}(x; \theta_t).W^{t})$ using the cross-entropy loss as in Equations~\ref{eq:1} and~\ref{eq:2}.

The teacher model is fine-tuned end-to-end with the labeled data to learn the parameters $\tilde{\theta^t}$.

\section{Distillation Features}
After fine-tuning the teacher for the given task, we can leverage representations learned by it to guide the student. To this end, we use different kinds of information from the teacher.

\subsection{Teacher Logits}

\subsubsection{Soft Predictions}
The cross-entropy loss for classification in Equation~\ref{eq:2} is computed over hard labels (e.g., binary 0/1). Training a student model on this objective allows it to learn the easy targets. Prior works on knowledge distillation in the vision community~\citep{DBLP:conf/nips/BaC14,DBLP:journals/corr/HintonVD15} show that the {\em logits} on the other hand, allow the student to train on the difficult targets. 

Logits are logarithms of predicted probabilities $logit(p) = log \frac{p}{1-p}$. The logits provide a better view of the teacher by emphasizing on the different relationships learned by it across different instances. For example, the review ``I loved the movie" has a small probability of being negative; whereas, the review ``The movie could have been better" can be positive or negative depending on context. Such uncertainties are reflected in the classification probabilities that provide better training targets than just the binary label as positive or negative. Furthermore, for multi-class classification with a lot of classes, the student model benefits from the non-zero targets spread over all the classes by the teacher rather than training on one-hot targets from the ground-truth labels.

Consider $p^t(x)$ to be the classification probabilities of an instance $x$ generated by the fine-tuned teacher model with $logit(p^t(x))$ representing the corresponding logits. Our objective is to train a student model with these logits as targets. 

Consider $z^s(x)$ to be the representation generated by the student for an instance $x$. We can use this representation to generate the classification scores for the instance by linear regression (since the targets are logits). Therefore, the classification scores for the instance is given by $r^s(x) = W^r \cdot z^s(x) + b^r$ where $W^r \in R^{C \cdot |z^s|}$ and $b^r \in R^C$ are trainable parameters and $C$ is the number of classes.

Now, we want to train the student neural network end-to-end by optimizing the following:

\begin{equation}
\label{eq:3}
    \min_{\theta^s, W^r, b^r} \mathbb{E}_{(x_u) \in \mathcal{D}_u} \mathcal{L_{LL}}({r}^s(x_u;\theta_s, W^r, b^r), logit(p^t(x_u);\tilde{\theta}_t))
\end{equation}

\noindent where $\mathcal{L_{LL}}$ denotes the loss between the classification scores generated by the student and the target logits generated by the teacher. The loss is given by the element-wise mean-squared error as:

\begin{equation}
\label{eq:4}
    \mathcal{L_{LL}} = \frac{1}{2}\sum_{x_u \in D_u} ||r^s(x_u;\theta_s, W^r, b^r)- logit(p^t(x_u;\tilde{\theta_t}))||^2
\end{equation}

\subsubsection{Hard Predictions} 

Instead of soft logits from the teacher, we can also binarize them by considering $argmax_{c \in C}\ p_c^t(x)$ to train the student. As we will see, this is one of the strongest baselines given a sufficiently powerful teacher.   

\subsection{Hidden Teacher Representations}

Different layers of a deep neural network abstract different kinds of representation from the input feature space with the top ones providing task-specific representation. Therefore, we can use the intermediate representation as learned by the teacher as a signal to aid the student in the learning process.

The primary objective of distillation is to train a student to mimic the teacher. The best student should generate very similar representation as the teacher. Since we want to perform task-specific distillation, we choose the representation generated by the last layer of the teacher that contains the most task-specific information. Therefore, the student is trained to predict the output of the teacher's last layer.

Consider $z^s(x; \theta_s)$ and $z^t(x; \tilde{\theta_t})$ to be the representations generated by the student and the fine-tuned teacher respectively for a given input sequence $x$. Consider $x_u \in D_u$ to be the set of unlabeled instances.

Note that the dimension of the representation generated by the student and the teacher could be different as they have very different architectures in our setting. In order to make the dimensions compatible for our optimization objective, we perform a non-linear transformation on the representation $z^s$ generated by the student as: $\tilde{z}^s(x)=gelu(W^f \cdot z^s(x) + b^f)$, where $W^f \in R^{|z^t| \cdot |z^s|}$ is the transformation matrix, $b^f \in R^{|z^t|}$ is the bias, and {\em gelu} (Gaussian Error Linear Unit)~\citep{DBLP:journals/corr/HendrycksG16} is the non-linear activation function. After the transformation, the student representation has the same dimension as that of the teacher's representation from their last layers. We can train the student neural network by optimizing the following:

\begin{equation}
\label{eq:5}
    \min_{\theta^s, W^f, b^f} \mathbb{E}_{x_u \in \mathcal{D}_u} \mathcal{L}_{RL}(\tilde{z}^s(x_u;\theta_s, W^f, b^f), z^t(x_u;\tilde{\theta}_t))
\end{equation}

\noindent where $\mathcal{L_{RL}}$ denotes the loss between the student and the teacher for the representations generated by their last layer. To this end, we compute the element-wise mean-squared error (mse) between the hidden state representations that is shown to perform better for distillation than other loss functions like the {\em KL}-divergence~\citep{DBLP:conf/nips/BaC14}:

\begin{equation}
\label{eq:6}
    \mathcal{L_{RL}} = \frac{1}{2}\sum_{x_u \in D_u} ||\tilde{z}^s(x_u;\theta_s, W^f, b^f)- z^t(x_u;\tilde{\theta_t})||^2
\end{equation}

\begin{figure}
\centering
  \includegraphics[width=0.7\linewidth]{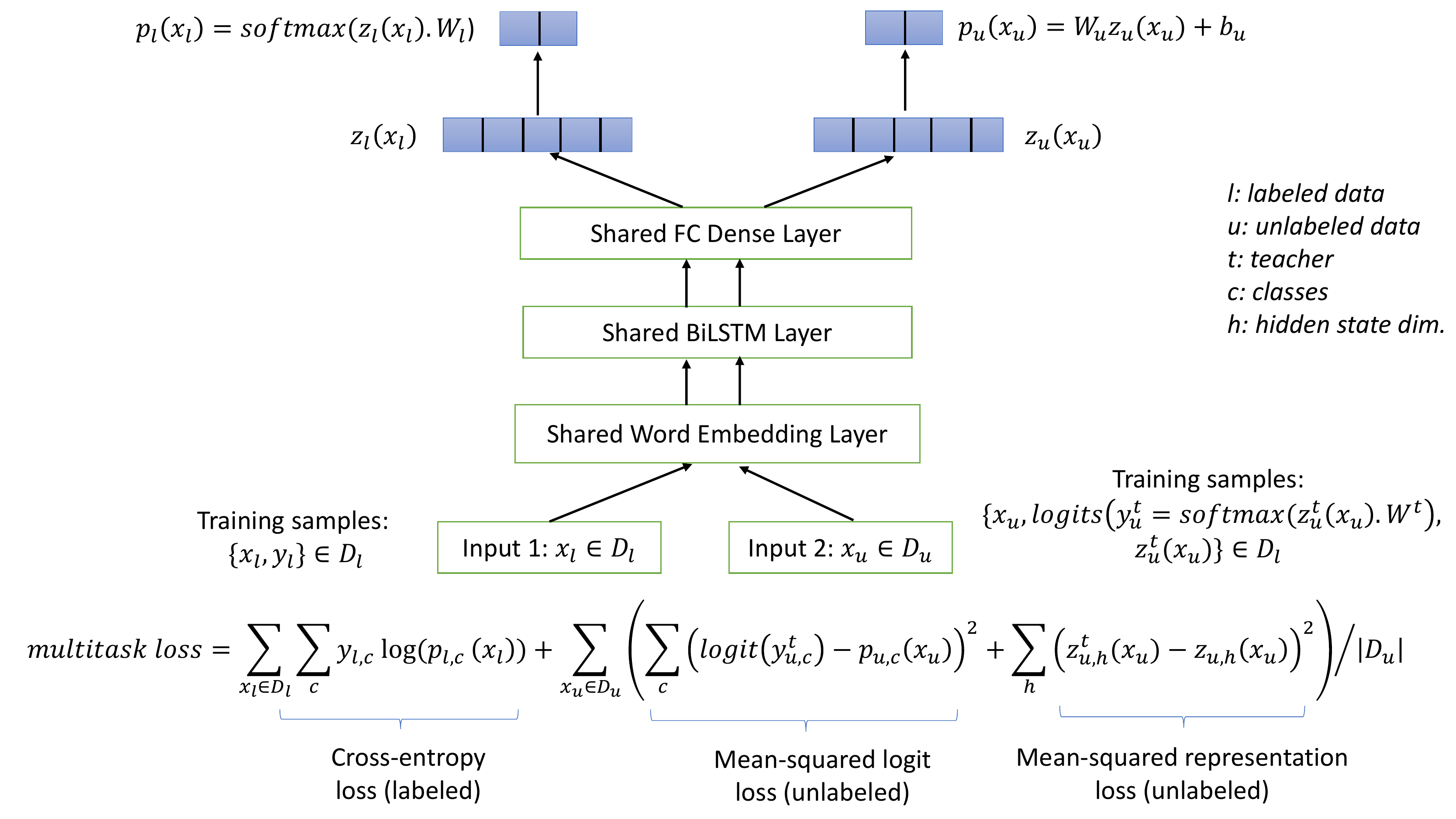}
  \caption{Multi-task student teacher distillation architecture.}
  \label{fig:arch}
\end{figure}

\section{Training}

In order to train the student model end-to-end, we want to optimize all the three loss functions $\mathcal{L_{CE}}, \mathcal{L_{RL}}$ and $\mathcal{L_{LL}}$. We can schedule these optimizations differently, and accordingly obtain different training regimens as follows:

\subsection{Joint Optimization}

In this, we optimize the following loss functions jointly:

\begin{multline}
    \frac{1}{|D_l|} \sum_{\{x_l, y_l\}\in D_l} \alpha \cdot \mathcal{L_{CE}}(x_l, y_l) +  \frac{1}{|D_u|} \sum_{\{x_u, y_u\}\in D_u} \bigg( \beta \cdot \mathcal{L_{RL}} (x_u, y_u) + \gamma \cdot \mathcal{L_{LL}} (x_u, y_u) \bigg)
\end{multline}

\noindent where $\alpha, \beta$ and $\gamma$ serve two functions: (i) scaling different loss functions to be in the same decimal range of magnitude and (ii) weighing the contribution of different losses. As discussed earlier, a high value of $\alpha$ makes the student model focus more on the easy targets; whereas a high value of $\gamma$ makes the student focus on the difficult targets as well as adapt the model to noisy ground-truth labels. The latter is not a focus of this work and omitted from further analysis.

The above loss function is computed over two different segments of the data. The first part involves cross-entropy loss computed over ground-truth training labels, whereas the second part involves representation and logit loss computed over unlabeled instances. The student model at each step of training receives two batches of data simultaneously from the data generator consisting of equivalent number of instances from the labeled data and the unlabeled ones. 

\subsection{Stagewise Training with Gradual Unfreezing}

Instead of optimizing all of the above loss functions jointly, we could adopt a stage-wise optimization scheme. In this, we first train the student neural network optimizing for the representation loss $\mathcal{R_{RL}}$. The model learns the parameters for the word embeddings, BiLSTM and the non-linear transformation functions for the final layer. This stage allows the student model to learn parameters so as to mimic the representations from the last layer of the teacher.

In the second stage, we optimize for the cross-entropy loss $\mathcal{R_{CE}}$ and the logit loss $\mathcal{R_{LL}}$ for the given task. However, rather than optimizing all the parameters all at once which risks `catastrophic forgetting'~\citep{DBLP:conf/acl/RuderH18}, we adopt a different approach. We gradually unfreeze the layers starting from the top layer that contains the most task-specific information and therefore allows the model to configure the task-specific layer first while others are frozen. In the next phase, the latter layers are unfrozen one by one and the model trained till convergence. Note that once a layer is unfrozen, it maintains the state. When the last layer (word embeddings) is unfrozen, the entire network is trained end-to-end till convergence.

\subsection{Stagewise Training: Distil Once Fine-tune Forever}

This is similar to the above setting with a tweak. In the first stage of optimization, the student model is trained end-to-end to jointly optimize the representation loss $\mathcal{R_{RL}}$ and the logit loss $\mathcal{R_{LL}}$ over the unlabeled data without requiring any access to ground-truth training labels.

In the second stage, we use the now trained student model to optimize for cross-entropy loss $\mathcal{R_{CE}}$ over labeled data for the given task and fine-tune its parameters. While doing so, we follow the same gradual unfreezing approach as above. 

Note that the first stage produces a distilled student model. We can now fine-tune it forever based on labeled data as more and more of it is available over time.

\section{Experiments}

\subsubsection{Dataset Description}
We perform large-scale experiments with data from four domains for different tasks as summarized in Table~\ref{tab:dataset}. Two of these datasets IMDB~\citep{DBLP:conf/acl/MaasDPHNP11} and Elec~\citep{DBLP:conf/recsys/McAuleyL13} are used for sentiment classification for movie reviews and Amazon electronics product reviews respectively. The other two datasets DbPedia~\citep{DBLP:conf/nips/ZhangZL15} and Ag News~\citep{DBLP:conf/nips/ZhangZL15} are used for topic classification of Wikipedia and news articles respectively. These datasets have been extensively used for benchmarking in several other text classification tasks~\citep{DBLP:conf/nips/ZhangZL15,DBLP:journals/corr/Johnson014,DBLP:conf/iclr/MiyatoDG17}

\begin{table}[]
    \centering
    \begin{tabular}{lccccc}
        \toprule
         { Dataset} & { Class} & { Train} & { Test} & { Unlabeled} & { \#W}\\\midrule
         IMDB & 2 & 25K & 25K & 50K & 235\\
         DBPedia & 14 & 560K & 70K & - & 51\\
         AG News & 4 & 120K & 7.6K & - & 40\\
         Elec & 2 & 25K & 25K & 200K & 108\\
         \bottomrule
    \end{tabular}
    \caption{Full dataset summary (W: average words per doc).}
    \label{tab:dataset}
\end{table}

\subsubsection{Evaluation Questions} In the experiments we want to answer the following questions:

\begin{itemize}
    \item How much can we improve the performance of student models by distillation?
    \item Does distillation improve with a better teacher?
    \item What is the impact of soft vs. hard distillation for pre-trained language models?
    \item When does distillation provide the maximum value?
\end{itemize}

\subsubsection{Parameter Configurations}

We split each labeled dataset into $90\%$ for training and $10\%$ for validation and report aggregate results on the blind test set based on the validation set. We implement our framework in Tensorflow and use four Tesla V100 gpus for experimentation. We use Adadelta~\citep{DBLP:journals/corr/abs-1212-5701} as the optimization algorithm with early stopping and use the best model found so far from the validation loss. Adadelta does not require an initial learning rate. We also experimented with Adam~\citep{DBLP:journals/corr/KingmaB14}. We found it to converge faster but the results were often sub-optimal compared to Adadelta in our setting. 
To regularize our network, we used dropouts ($rate=0.4$) after every layer as well as recurrent dropouts ($rate=0.2$) in the BiLSTM. We use pre-trained $300$ dimensional word embeddings from Glove~\citep{DBLP:conf/emnlp/PenningtonSM14} to initialize our model that are fine-tuned during training. Words not present in the vocabulary are randomly initialized with $\mathcal{U}(-.1, .1)$. We set the number of hidden units in the LSTM to $600$ for each direction ($1200$ for BiLSTM), batch size to 64, and $\alpha=\beta=10, \gamma=1$.

\subsubsection{Teacher Model}

We consider the state-of-the-art pre-trained language model BERT~\citep{DBLP:conf/naacl/DevlinCLT19} to be the teacher in our setting. Note that we do not make any assumptions about the architecture of the teacher; it is possible to plug-in any large pre-trained model in our framework.

BERT's model architecture is a multi-layer bidirectional Transformer encoder. We experiment with two configurations: (i) BERT Base consisting of 12 layers, 12 self-attention heads, and 768 dimensional hidden state representation with an overall 110 million parameters. (ii) BERT-Large consisting of 24 layers, 16 self-attention heads, and 1024 dimensional hidden state representation with an overall 340 million parameters. Pre-trained BERT Large models have two variants corresponding to different masking technique for pre-training. We use uncased, whole word masking models that are shown to be better than previous versions.

\subsubsection{Student Model}

We compare the following models:

\noindent (i) {\bf RNN Encoder} We consider BiLSTM encoders with word embeddings. The last hidden state of BiLSTM is fed into softmax for classification and the network parameters are trained by optimizing cross-entropy loss over labeled data as in Equations~\ref{eq:1} and~\ref{eq:2}. We use a basic tokenizer with this model that lowercases all words and splits by whitespace.

\noindent (ii) {\bf Student Model without Distillation} This extends the previous one with a max-pooling mechanism to consider all the BiLSTM hidden states instead of just the last one, while still generating a fixed length vector. Also, this and the subsequent models use wordpiece tokenization to normalize the input text. The student model is trained over labeled data using cross-entropy loss similar to the RNN encoder.

\noindent (iii) {\bf Student Model with Distillation} 
In this, we distil the aforementioned student with (soft / hard) targets and representations from the teacher. First, we fine-tune the teacher on labeled data and use it to generate the logits and hidden state representations for unlabeled instances. We train the student model end-to-end using cross-entropy loss on labeled instances (Equations~\ref{eq:1},~\ref{eq:2}) as well as logit loss (Equations~\ref{eq:3},~\ref{eq:4}) and representation loss (Equations~\ref{eq:5},~\ref{eq:6}) on the unlabeled data. We test three different learning strategies based on a joint optimization scheme as well as two stagewise ones with gradual unfreezing of the intermediate layers.


\subsection{Distillation with BERT Base}

In this setting, we consider BERT Base as the teacher with $110M$ parameters. For our setup, we require both labeled instances and unlabeled ones. However, some of the datasets like Dbpedia and AG News do not have any unlabeled instances. Therefore, we split all the labeled instances into two parts as follows: (a) randomly sample equal number of instances per class from the labeled data and use them for training as $D_L$. The total size of the training data for each task is made equal to the size of the test data. (b) all the remaining instances from the labeled data are stripped of labels and considered as unlabeled instances $D_U$ for our framework. Table~\ref{tab:moderate} shows the derived dataset statistics. Unless, otherwise mentioned all the experimental results in this paper are reported on this data split.


\begin{table}[]
    \centering
    \begin{tabular}{lrrrr}
        \toprule
         { Dataset} & { Train} & { Test} & { Unlabeled} & { \#Train/Class} \\\midrule
         IMDB & 25K & 25K & 50K & 12.5K \\
         DBPedia & 70K & 70K & 490K & 5k\\
         AG News & 7.6K & 7.6K & 112.4K & 1.9K\\
         Elec & 25K & 25K & 200K & 12.5K\\
         \bottomrule
    \end{tabular}
    \caption{Derived dataset summary.}
    \label{tab:moderate}
\end{table}

In this experiment, we compare the performance of the different student models with our distilled model. Table~\ref{tab:distil} summarizes the overall results. 

\begin{table}
    \centering
    \begin{tabular}{lcccc}
    \toprule
{ Dataset} & { RNN} & { Student} & { Student } & { BERT}\\
& & { no distil.} & { with distil.} & Base \\
\midrule
Ag News &	87.66 &	89.71 &	{\bf 92.33} &	92.12\\
IMDB &	89.93 &	89.37 &	91.22 &	{\bf 91.70}\\
Elec &	89.72 &	90.62 &	{\bf 93.55} &	{93.46}\\
Dbpedia	& 98.44	& 98.64	& 99.10	& {\bf 99.26}\\
\bottomrule
    \end{tabular}
    \caption{Distillation performance with BERT Base.}
    \label{tab:distil}
\end{table}

We observe that the student model with the max-pooling mechanism and wordpiece tokenization performs slightly better than the RNN encoder without these mechanisms. We observe that the RNN in general performs quite well in all the tasks for all of the datasets. With distillation we are further able to improve the student performance by $2$ percent on an average over all the tasks and match the teacher performance. From Table~\ref{tab:param} we observe that the distilled student model is $8.5x$ smaller in size than the teacher model in terms of the number of parameters. 

\begin{table}[h]
    \centering
    \begin{tabular}{lcccc}
    \toprule
{ Dataset} & { Distil with} & { Distil with} &  {\small BERT} & {\small	BERT}\\
 & { {\small BERT Base}} & { {\small BERT Large}} & {\small Base} & {\small Large} \\
\midrule
Ag News &	92.33 &	94.33 & 92.12	& {\bf 94.63}\\
IMDB &	91.22 &	91.70 &	91.70 &	{\bf 93.22}\\
Elec & 93.55 &	93.56 &	93.46 &	{\bf 94.27}\\
DbPedia	& 99.10	& 99.06 &	{\bf 99.26}	& 99.20\\
\bottomrule
    \end{tabular}
    \caption{Better distillation with better teacher.}
    \label{tab:teacher}
\end{table}

\subsection{Distillation with Larger and Better Teacher}

In this, we want to find out if the distillation performance improves with a more powerful teacher. We therefore experiment with BERT Large that is $3x$ larger than the Base version while  performing better in many of the tasks~\citep{DBLP:conf/naacl/DevlinCLT19}. Table~\ref{tab:teacher} shows the performance comparison. We observe that the student model performance improves marginally as the performance of the teacher improves. However, the gap in performance between the teacher and student also increases compared to the Base version indicating saturation in the the student network.

\subsubsection{Parameters} Table~\ref{tab:param} shows the number of parameters for the different models. Due to the usage of wordpiece tokenization with $30K$ fixed vocabulary size, all the model sizes have a fixed upper-bound irrespective of the dataset. We show the average number of parameters for the student model over different settings and configurations. Distilling the student with a larger teacher only increases the dimensionality of the transformation matrix $W^f$ to align the last hidden state representation of the teacher and student.

\begin{table}[h]
    \centering
    \begin{tabular}{ccc}
    \toprule
    Distilled Student & BERT Base & BERT Large  \\\midrule
    13M & 110M & 340M\\\bottomrule
    \end{tabular}
    \caption{Parameters for the different models (M: millions).}
    \label{tab:param}
\end{table}

\subsection{Distilling Hard Targets vs. Soft Logits}

In the above experiments, we found the distillation performance with BERT Large to be better than that with BERT Base. Therefore, in this setting, we consider BERT Large as the teacher fine-tuned on labeled data. We use the fine-tuned teacher model to generate {\em hard predictions} on the {\em unlabeled data}. Thereafter, we train our student model jointly on the originally labeled data and BERT Large labeled instances.

Table~\ref{tab:hard} shows the distillation performance with hard targets versus leveraging the soft predictions and hidden state representations. We observe that the student model trained on hard BERT Large predictions perform extremely well. With soft distillation, we are able to further improve the performance across all the tasks.

\begin{table}[]
    \centering
    \begin{tabular}{lcc}
    \toprule
    Dataset & Soft Distillation & Hard Distillation  \\
    \midrule
    AG News & {\bf 94.33} & 93.17\\ 
    Elec & {\bf 93.56} & 92.57 \\
    DBPedia & {\bf 99.06} & 98.97 \\
    IMDB & {\bf 91.70} & 91.21\\
    \bottomrule
    \end{tabular}
    \caption{Soft vs. hard distillation results with BERT Large.}
    \label{tab:hard}
\end{table}

However, we observe the performance gap between soft distillation and hard targets from BERT to be narrow. To investigate this, we compute the variance in prediction probabilities of an instance from the teacher computed as follows. Consider $p_c(x_i)$ to be the probability of $x_i$ belonging to class $c$ as predicted by the teacher. Overall variance in prediction is given by $\sum_{i=1}^N Var(\langle p_1(x_i), p_2(x_i), \cdots p_c(x_i) \rangle) / N$ aggregated over all instances. Maximum variance is obtained when exactly one of these elements is one and rest are zeros.

Table~\ref{tab:variance} shows the prediction variance of the teacher for different tasks. The table also shows the maximum variance possible in case of one-hot predictions. We observe that BERT predictions are extremely confident almost similar to hard labels. The variance and therefore, the confidence further increases with the size of the teacher model. Since soft distillation relies on exploiting teacher uncertainties to improve the student -- such confident teacher makes this process extremely difficult. Nevertheless, we are still able to regularize the model with soft distillation and further improve the performance over that of hard targets from teacher.

\begin{table}[]
    \centering
    \begin{tabular}{lcccc}
    \toprule
Dataset &	BERT & BERT & Max. & Classes\\
 & Base & Large & Var. &\\
 \midrule
IMDB & 0.240 & 0.249 & 0.250 & 2 \\
Elec & 0.239 & 0.247 & 0.250 & 2\\
Ag News	& 0.173 & 0.185 & 0.188 & 4\\
Dbpedia &	0.066(0) &	0.066(1) & 0.066(3) & 14\\
\bottomrule
    \end{tabular}
    \caption{Variance in prediction probabilities of the teacher. Maximum variance is obtained when the teacher spits hard labels as opposed to soft predictions.}
    \label{tab:variance}
\end{table}

\subsection{ Distillation with Less Training Labels}

\subsubsection{$500$ labeled samples per class} In the previous experiments, we observe that the distillation process is able to improve the performance of the student model when trained over thousands of labeled instances per class. In this setting, we consider only $500$ labeled samples per class for every task, while the remaining instances are considered as unlabeled in our framework. The objective of this experiment is to find out scenarios where distillation with unlabeled data provides maximum value. Table~\ref{tab:less} summarizes the results.

\begin{table}[]
    \centering
    \begin{tabular}{lcccc}
\toprule
{ Dataset} & { RNN} & { Student} & { Student}  &	{ BERT}\\
 &  & { no distil.} & { with distil.}  &	{ Large}\\
\midrule
AG News & 85.03 &	85.85 &	{\bf 90.45} &	90.36\\
IMDB & 52.94 &	61.53 &	89.08 &	{\bf 89.11}\\
Elec & 68.93 &	65.68	& {\bf 91.00}	& 90.41\\
DBpedia & 93.33 &	96.30 &	{\bf 98.94}	& {\bf 98.94}\\
\bottomrule
    \end{tabular}
    \caption{Distillation with BERT Large on $500$ labeled samples per class.}
    \label{tab:less}
\end{table}

We observe a huge improvement in performance on distilling the student model from BERT Large when there are limited number of labeled samples per class. The distilled student improves over the non-distilled version by $19.4$ percent and matches the teacher performance for all of the tasks.

\begin{table}[]
    \centering
    \begin{tabular}{lccc}
    \toprule
    Dataset & Student & Student & BERT \\
    & Hard Distil. & Soft Distil. & Large \\\midrule
    IMDB &  87.37 & {\bf 87.84} & 87.72\\
    Elec  & 89.17 & {\bf 90.14} & 89.94\\
        \bottomrule
    \end{tabular}
    \caption{Distillation with BERT Large (with pre-training and fine-tuning) on $100$ labeled samples per class.}
    \label{tab:small}
\end{table}

\subsubsection{$100$ labeled samples per class} In this setting, we consider only $100$ labeled samples per class for training all models. When we fine-tuned BERT Large on this small labeled set, the accuracy was similar to the previous case ($500$ labeled samples per class) for DbPedia and AG News dataset. However, for IMDB and Elec datasets BERT accuracy was $50\%$ (random) for binary classification. Therefore, we now turn our focus to these two challenging datasets for this setting.

First, we pre-train the BERT Large language model on these two datasets starting from the checkpoints. Second, we fine-tune the model for classification tasks. Note that for all of the previous experiments, we only performed the second step and never pre-trained the language models. Surprisingly, its performance now improved drastically. Next, we distil the student model from the fine-tuned BERT Large classifiers. Table~\ref{tab:small} summarizes the results. We observe that the student model now exceeds the BERT Large performance with just $100$ labeled samples per class. In contrast to the previous settings where we did not perform the language model pre-training, the student model is now able to exploit the soft representations better from the teacher.


\subsection{Impact of Training Schedule on Distillation}

We had proposed three different schedules for distilling the student model based on joint and stagewise optimization with gradual unfreezing. From Table~\ref{tab:training} we observe that the different training schedules does impact the distillation performance. We observe that the stagewise schedule ({\small $\mathcal{R_{RL}}\ | \mathcal{R_{LL}} + \mathcal{R_{CE}}$} in table) where we first optimize the loss $\mathcal{R_{RL}}$ (cf. Equations~\ref{eq:5},~\ref{eq:6}) between the representations of the student and teacher model from the final layer, and then jointly optimize for the cross-entropy loss $\mathcal{R_{CE}}$ (cf. Equations~\ref{eq:1},~\ref{eq:2}) and logit loss $\mathcal{R_{LL}}$ (cf. Equations~\ref{eq:3},~\ref{eq:4}) with gradual unfreezing performs the best.

\begin{table}[]
    \centering
    \begin{tabular}{lccc}
\toprule
Dataset & {\small $\mathcal{R_{RL}}\ |$} & {\small $\mathcal{R_{RL}} +$} &
{\small $\mathcal{R_{RL}}+ \mathcal{R_{LL}}\ |$}\\
 & {\small $\mathcal{R_{LL}} + \mathcal{R_{CE}}$} & {\small $\mathcal{R_{LL}} + \mathcal{R_{CE}}$} &
{\small $\mathcal{R_{CE}}$}\\
\midrule
Ag News	& {\bf 92.33} &	92.17 &	92.26\\
IMDB & {\bf 91.22} &	90.07 & 90.87\\
Elec & {\bf 93.55} &	{93.12} & 93.10\\
DbPedia	 & {\bf 99.10} & {99.08} & {99.08}\\
\midrule
Ag News	& {\bf 94.33} &	93.90 &	93.78\\
IMDB & {\bf 91.70} &	90.84 & 90.89\\
Elec & {\bf 93.56} &	93.16 & 93.35\\
DbPedia	 & {\bf 99.06} & {99.06} & 99.04 \\
\bottomrule
    \end{tabular}
    \caption{Different training schedules for distillation with BERT. `$|$' indicates different stages and `$+$' indicates joint optimization. The first block shows distillation results with BERT Base and the second block with BERT Large. The training scheme in the first column optimizes the representation loss in the first stage, and then jointly optimizes the logit loss and cross-entropy loss in the second stage with gradual unfreezing of the layers. The one in the second column optimizes all the loss functions jointly. Whereas the last one first optimizes the representation and logit loss over unlabeled data, followed by fine-tuning on labeled data.}
    \label{tab:training}
\end{table}

\subsection{Summary of Findings}
In this work, we make the following observations:

\noindent $\bullet$  RNN encoders (with word embeddings) work very well for distilling large pre-trained language models. We also experimented with encoders like deep multi-layer CNN as in~\citep{DBLP:journals/corr/RomeroBKCGB14,DBLP:conf/nips/ZhangZL15}. However RNNs performed much better.

\noindent $\bullet$  Several architectural choices like width of the student network, sequence length, regularization (e.g., dropouts) and optimization algorithm impact distillation performance.

\noindent $\bullet$  Students with distillation greatly outperform those without in settings with limited amount of labeled data, large amount of in-domain unlabeled data, and a good teacher.

\noindent $\bullet$  Large pre-trained language models make very confident predictions. This makes soft distillation difficult which is more effective in presence of teacher uncertainties and possibly a large label space.

\noindent $\bullet$  Increasing the size of the teacher leads to marginal gain in the distillation process without increasing the size of the student network once it saturates.

\noindent $\bullet$  Training schedule for optimizing multiple loss functions impact distillation where stagewise training with gradual unfreezing works best.

\section{Related Work}

\noindent{\bf Pre-trained language models:} Deep neural networks have been shown to perform extremely well for several natural language processing tasks including text classification~\citep{DBLP:conf/nips/ZhangZL15,DBLP:journals/corr/Johnson014,DBLP:conf/iclr/MiyatoDG17}. However these models require lots of labeled training data. Recent advances in pre-training help in this regard. Large-scale pre-trained language models such as BERT~\citep{DBLP:conf/naacl/DevlinCLT19}, GPT~\citep{radford2018improving,radford2019}, ELMO~\citep{DBLP:conf/naacl/PetersNIGCLZ18} and XLNet~\citep{DBLP:journals/corr/abs-1906-08237} have been developed that are trained in a self-supervised fashion employing language model objectives. These models benefit from large amounts of unlabeled text and increasingly complex networks with millions to billions of
parameters. Although these pre-trained models are the state-of-the-art for several NLP tasks, a big challenge is to deploy them in practise.

\noindent{\bf Model compression and knowledge distillation:} Prior works in the vision community dealing with huge architectures like AlexNet and ResNet have addressed this challenge in two ways. Works in model compression use quantization~\citep{DBLP:journals/corr/GongLYB14}, low-precision training and pruning the network, as well as their combination~\citep{HanMao16} to reduce the memory footprint. On the other hand, works in knowledge distillation leverage student teacher models. In this a shallow student model is trained to mimic the huge teacher model. These approaches include using soft logits as targets~\citep{DBLP:conf/nips/BaC14}, increasing the temperature of the softmax to match that of the teacher~\citep{DBLP:journals/corr/HintonVD15} as well as using internal teacher representations~\citep{DBLP:journals/corr/RomeroBKCGB14} 
(refer to~\citep{DBLP:journals/corr/abs-1710-09282} for a survey). 

Very recent works in NLP have focused on multi-task learning to distil knowledge from multiple big learners into a single big learner~\citep{DBLP:journals/corr/abs-1904-09482} and using heuristic rule based data augmentation for distillation~\citep{DBLP:journals/corr/abs-1903-12136}. In contrast to these, we study several aspects of distillation and show smaller student models can match the performance of a huge accurate teacher given large unlabeled transfer data. Refer to introduction for an overview of other concurrent works.

\section{Conclusions}

In this work, we perform an extensive study of knowledge distillation from large pre-trained language models. We explore several aspects like distillation objective, representations to harness from teacher, training schedule and impact of labeled instances. We show that RNN encoders like BiLSTMs (as students) can perform at par with large pre-trained language models like BERT (as teachers) with greatly reduced parameter space (9x-26x) given large amount of unlabeled transfer data with matching performance and sometimes marginally exceeding the teacher. Additionally, we show that distillation empowers students more in presence of limited amount of labeled training data.
\bibliographystyle{plainnat}
\bibliography{distil}
\end{document}